\documentclass[10pt,twocolumn,letterpaper]{article}

\usepackage[pagenumbers]{cvpr} 

\usepackage[dvipsnames, table]{xcolor}
\definecolor{mblue}{HTML}{0055CC}
\definecolor{mred}{HTML}{CC1100}
\definecolor{mpurple}{HTML}{A8D08D}

\usepackage{microtype}                                  
\usepackage{amsmath, amsthm, amssymb, amsfonts}
\usepackage{makecell}
\usepackage{adjustbox}
\usepackage{multirow}
\usepackage{pifont}

\newcommand{\mtab}{Table~}                              

\newcommand{\mpara}{\noindent\textbf}

\newcommand{\mce}{\mathcal{E}}
\newcommand{\mcf}{\mathcal{F}}
\newcommand{\mcg}{\mathcal{G}}

\newcommand{\mcj}{\mathcal{J}}

\newcommand{\mcv}{\mathcal{V}}

\newcommand{\mcx}{\mathcal{X}}

\newcommand{\mcz}{\mathcal{Z}}

\newcommand{\lbr}{\left(}
\newcommand{\rbr}{\right)}

\newcommand{\lbe}{\left\lbrace}
\newcommand{\rbe}{\right\rbrace}

\definecolor{mgray}{HTML}{DDDDDD}
\definecolor{mred}{HTML}{F38181}

\usepackage[pagebackref,breaklinks,colorlinks,linkcolor=mred,citecolor=mblue]{hyperref}  
\begin{document}

\title{SEGT: A General Spatial Expansion Group Transformer for nuScenes Lidar-based Object Detection Task}

\author{
  Cheng Mei\textsuperscript{$1$} \quad 
  Hao He\textsuperscript{$1$} \quad
  Yahui Liu\textsuperscript{$1$} \quad
  Zhenhua Guo\textsuperscript{$1$}\thanks{Corresponding Author} \quad
  \vspace{0.1cm} \\
  \textsuperscript{$1$}Tianyijiaotong Technology Ltd.
  \vspace{0.1cm} \\
  \texttt{\{cheng.mei, hao.he, yahui.liu zhenhua.guo\}@tyjt-ai.com}
}

\maketitle
\begin{abstract}
In the technical report, we present a novel transformer-based framework for nuScenes lidar-based object detection task, termed Spatial Expansion Group Transformer (SEGT). To efficiently handle the irregular and sparse nature of point cloud, we propose migrating the voxels into distinct specialized ordered fields with the general spatial expansion strategies, and employ group attention mechanisms to extract the exclusive feature maps within each field. Subsequently, we integrate the feature representations across different ordered fields by alternately applying diverse expansion strategies, thereby enhancing the model's ability to capture comprehensive spatial information. The method was evaluated on the nuScenes lidar-based object detection test dataset, achieving an NDS score of 73.9 without Test-Time Augmentation (TTA) and 74.5 with TTA, demonstrating the effectiveness of the proposed method. Notably, our method ranks the 1st place in the nuScenes lidar-based object detection task.
\end{abstract}

\vspace{-1em}
\section{Introduction}
\label{sec:introduction}
Lidar-based object detection has become a critical task in the fields of autonomous driving and robotics, due to its ability to capture rich three-dimensional spatial information in diverse environmental conditions. The nuScenes dataset, a comprehensive and large-scale benchmark designed specifically for autonomous vehicle perception, presents significant challenges for lidar-based object detection, primarily due to the inherent sparsity and irregularity of lidar point clouds. While conventional methods have demonstrated some success, they often struggle to fully exploit the complex spatial structure of lidar data, especially in terms of handling variations in point cloud density and the irregular distribution of points.

In this technical report, we introduce a novel transformer-based framework designed to address these challenges, called the Spatial Expansion Group Transformer (SEGT). Our approach aims to efficiently process the irregular and sparse nature of point clouds by proposing a novel technique to organize the data. Specifically, we migrate the lidar points into distinct specialized ordered fields using general spatial expansion strategies. This migration allows the model to focus on spatial regions of interest and handle variations in point density more effectively. To further enhance the model’s performance, we apply group attention mechanisms to capture and extract exclusive feature maps within each field, ensuring that each region's unique spatial characteristics are preserved and emphasized. Additionally, we integrate these feature representations across different ordered fields by alternately applying diverse spatial expansion strategies, thereby enhancing the model’s ability to capture comprehensive spatial information.

We evaluate the performance of SEGT on the nuScenes lidar-based object detection test dataset. The results show that our method achieves an impressive NDS score of 73.9 without Test-Time Augmentation (TTA), and further improves to 74.5 with TTA. The results demonstrate the effectiveness of SEGT in handling the complexities of lidar-based object detection.

\vspace{-0.5em}
\section{Method}
\label{sec:method}
\begin{table*}[t!]
    \centering
    \vspace{-1em}
    \caption{The LiDAR-only non-ensemble 3D detection performance comparison on the nuScenes test set. The table is mainly sorted by nuScenes detection score (NDS) which is the official ranking metric. 'T.L.', 'C.V.', 'Ped.', 'M.T.', 'Byc.', 'T.C.', and 'B.R.' are short for trailer, construction vehicle, pedestrian, motor, bicycle, traffic cone, and barrier, respectively. $\ddagger$ means using test-time augmentation (TTA). All models listed take LIDAR data as input without image fusion or any model ensemble.} 
    \label{tab:ComparisonOnNuScenes}
    \begin{adjustbox}{width=1\linewidth}
        \begin{tabular}{c|cc|cccccccccc}
            \toprule
            Methods & NDS $\uparrow$ & mAP $\uparrow$ & Car & Truck & Bus & T.L. & C.V. & Ped. & M.T. & Byc. & T.C. & B.R. \\
            \midrule
            PointPillars~\cite{Lang2019PointPillars}                & 45.3 & 30.5 & 68.4 & 23.0 & 28.2 & 23.4 &  4.1 & 59.7 & 27.4 &  1.1 & 30.8 & 38.9 \\ 
            3DSSD~\cite{Yang20203DSSD}                              & 56.4 & 42.6 & 81.2 & 47.2 & 61.4 & 30.5 & 12.6 & 70.2 & 36.0 &  8.6 & 31.1 & 47.9 \\ 
            CBGS~\cite{Zhu2019GBGS}                                 & 63.0 & 52.8 & 81.1 & 48.5 & 54.9 & 42.9 & 10.5 & 80.1 & 51.5 & 22.3 & 70.9 & 65.7 \\ 
            CenterPoint~\cite{Yin2021CenterPoint}                   & 65.5 & 58.0 & 84.6 & 51.0 & 60.2 & 53.2 & 17.5 & 83.4 & 53.7 & 28.7 & 76.7 & 70.9 \\ 
            CVCNET~\cite{Chen2020CVCNet}                            & 66.6 & 58.2 & 82.6 & 49.5 & 59.4 & 51.1 & 16.2 & 83.0 & 61.8 & 38.8 & 69.7 & 69.7 \\ 
            HotSpotNet~\cite{Chen2020HotSpotNet}                    & 66.0 & 59.3 & 83.1 & 50.9 & 56.4 & 53.3 & 23.0 & 81.3 & 63.5 & 36.6 & 73.0 & 71.6 \\
            AFDetV2~\cite{Hu2022AFDetV2}                            & 68.5 & 62.4 & 86.3 & 54.2 & 62.5 & 58.9 & 26.7 & 85.8 & 63.8 & 34.3 & 80.1 & 71.0 \\ 
            FocalsConv~\cite{Chen2022FSC}                           & 70.0 & 63.8 & 86.7 & 56.3 & 67.7 & 59.5 & 23.8 & 87.5 & 64.5 & 36.3 & 81.4 & 74.1 \\ 
            MGTANet-P~\cite{Koh2023MGTANet}                         & 61.4 & 50.9 & 81.3 & 45.8 & 55.0 & 48.9 & 18.2 & 74.4 & 52.6 & 17.8 & 61.7 & 53.0 \\
            MGTANet-C~\cite{Koh2023MGTANet}                         & 71.2 & 65.4 & 87.7 & 56.9 & 64.6 & 59.0 & 28.5 & 86.4 & 72.7 & 47.9 & 83.8 & 65.9 \\
            VoxelNeXt~\cite{Chen2023VoxelNeXt}                      & 70.0 & 64.5 & 84.6 & 53.0 & 64.7 & 55.8 & 28.7 & 85.8 & 73.2 & 45.7 & 79.0 & 74.6 \\
            LargeKernel3D~\cite{Chen2023LargeKernel3D}              & 70.6 & 65.4 & 85.5 & 53.8 & 64.4 & 59.5 & 29.7 & 85.9 & 72.7 & 46.8 & 79.9 & 75.5 \\
            TransFusion-L~\cite{Bai2022Transfusion}                 & 70.2 & 65.5 & 86.2 & 56.7 & 66.3 & 58.8 & 28.2 & 86.1 & 68.3 & 44.2 & 82.0 & 78.2 \\ 
            DSVT~\cite{Wang2023DSVT}                                & 72.7 & 68.4 & 86.8 & 58.4 & 67.3 & 63.1 & 37.1 & 88.0 & 73.0 & 47.2 & 84.9 & 78.4 \\
            FSTR-L~\cite{Zhang2023FSTR}                             & 70.4 & 66.2 & 85.8 & 53.9 & 64.1 & 57.4 & 31.1 & 87.5 & 74.1 & 48.2 & 81.4 & 78.1 \\
            FSTR-XL~\cite{Zhang2023FSTR}                            & 71.5 & 67.2 & 86.5 & 54.1 & 66.4 & 58.5 & 33.4 & 88.6 & 73.7 & 48.1 & 84.4 & 78.1 \\
            HEDNet~\cite{Zhang2024HEDNet}                           & 72.0 & 67.7 & 87.1 & 56.5 & \textbf{70.4} & 63.5 & 33.6 & 87.9 & 70.4 & 44.8 & 85.1 & 78.1 \\
            FocalFormer3D~\cite{Chen2023FocalFormer3D}              & 72.6 & 68.7 & 87.2 & 57.0 & 69.6 & 64.9 & 34.4 & 88.2 & 76.2 & 49.6 & 82.3 & 77.8 \\
            Real-Aug~\cite{Zhan2023RealAug}                         & 70.9 & 65.8 & 85.2 & 54.6 & 65.2 & 60.0 & 31.3 & 85.7 & 72.6 & 46.4 & 80.0 & 77.0 \\
            LION~\cite{Liu2024LION}                                 & \textbf{73.9} & 69.8 & 87.2 & \textbf{61.1} & 68.9 & 65.0 & 36.3 & \textbf{90.0} & 74.0 & 49.2 & \textbf{87.3} & \textbf{79.5} \\
            \rowcolor[gray]{0.9}
            \textbf{SEGT}                                           & \textbf{73.9} & \textbf{70.1} & \textbf{87.3} & 60.7 & 69.3 & \textbf{66.7} & \textbf{37.8} & 88.5 & \textbf{77.2} & \textbf{49.7} & 85.7 & 77.8 \\
            \midrule
            CenterPoint$^{\ddagger}$~\cite{Yin2021CenterPoint}      & 67.3 & 60.3 & 85.2 & 53.5 & 63.6 & 56.0 & 20.0 & 84.6 & 59.5 & 30.7 & 78.4 & 71.1 \\
            MGTANet$^{\ddagger}$~\cite{Koh2023MGTANet}              & 72.7 & 67.5 & \textbf{88.5} & 59.8 & 67.2 & 61.5 & 30.6 & 87.3 & 75.8 & 52.5 & 85.5 & 66.3 \\
            LargeKernel3D$^{\ddagger}$~\cite{Chen2023LargeKernel3D} & 72.8 & 68.7 & 86.7 & 58.5 & 67.7 & 62.7 & 31.9 & 88.5 & 77.1 & 54.9 & 82.3 & 76.6 \\
            VoxelNeXt$^{\ddagger}$~\cite{Chen2023VoxelNeXt}         & 71.4 & 66.2 & 85.3 & 55.7 & 66.2 & 57.2 & 29.8 & 86.5 & 75.2 & 48.8 & 80.7 & 76.1 \\
            FSTR-L$^{\ddagger}$~\cite{Zhang2023FSTR}                & 72.9 & 69.5 & 87.2 & 57.0 & 69.3 & 61.0 & 35.7 & 89.8 & 77.9 & 52.8 & 84.0 & 79.8 \\
            FSTR-XL$^{\ddagger}$~\cite{Zhang2023FSTR}               & 73.6 & 70.2 & 87.1 & 58.4 & 69.3 & 60.6 & 32.9 & \textbf{90.0} & 80.2 & \textbf{57.0} & 85.7 & \textbf{80.6} \\
            FocalFormer3D$^{\ddagger}$~\cite{Chen2023FocalFormer3D} & 73.9 & 70.5 & 87.8 & 59.4 & \textbf{73.0} & \textbf{65.7} & 37.8 & \textbf{90.0} & 77.4 & 52.4 & 83.4 & 77.8 \\
            Real-Aug$^{\ddagger}$~\cite{Zhan2023RealAug}            & 74.4 & 70.2 & 86.8 & 59.3 & 70.1 & 65.6  & 35.5 & 89.0 & 78.3 & 55.1 & 84.5 & 77.6 \\
            \rowcolor[gray]{0.9}
            \textbf{SEGT}$^{\ddagger}$                              & \textbf{74.5} & \textbf{71.2} & 86.8 & \textbf{62.7} & 70.7 & 63.8 & \textbf{39.3} & 88.9 & \textbf{81.9} & 51.5 & \textbf{86.7} & 80.1 \\
            \bottomrule
        \end{tabular}
    \end{adjustbox}
\end{table*}

We address lidar-based object detection in three steps. Firstly, the voxel feature maps and coordinates are obtained by dynamic voxel feature encoding, a widely adopted approach for voxelizing point clouds on OpenPCDet \cite{OpenPCDet2020, Chen2022FSC}, to simplify data processing and reduce computational complexity. Then, we employ gengeral \textit{spatial expansion group transformer encoder}, termed \textit{SEGT} encoder, to enhance the voxel representations, enabling a comprehensive integration with the spatial features of their surrounding neighborhood. To generate the prediction boxes, we employ the Transfusion head \cite{Bai2022Transfusion} as the detection head, consistent with several existing methods \cite{Bai2022Transfusion, Wang2023DSVT}, and convert the output of SEGT encoder into BEV (Bird's Eye View) features for its input.

\paragraph{SEGT Encoder} To efficiently process irregular and sparse voxels, we propose migrating the voxels into distinct specialized ordered fields, and employ group attention mechanisms to extract the exclusive feature maps within each field.  Subsequently, we integrate the feature representations across different ordered fields by alternately applying diverse expansion strategies, thereby enhancing the model’s ability to capture comprehensive spatial information

Specifically, for any given voxel feature maps $\mcf$ and coordinates $\mcv$ in field $\mcx$, we propose the general conjugate Hilbert expansion strategies $\mcj_{+}^{\mcx \rightarrow \mcz}$ and $\mcj_{-}^{\mcx \rightarrow \mcz}$ to migrate these into different ordered fields. Then, the group attention mechanisms are applied to extract the exclusive feature maps to significantly reduce the computational complexity and memory overhead of attention. Consequently, each SEGT layer can be represented as:
\begin{equation}
\begin{aligned}
\mcf_{?}^{\mcz}, \mcv_{?}^{\mcz} &= \mcj _{?}^{\mcx \rightarrow \mcz} \lbr \mcf^{\mcx}, \mcv^{\mcx}, L_{glb}, L_{lcl}\rbr,\\
\hat{\mcf}_{?}^{\mcz} &= \mcg \lbr \mcf_{?}^{\mcz}, \mce \lbr \mcv_{?}^{\mcz} \rbr, G \rbr,\\
\tilde{\mcf}^{\mcx} &= \tilde{\mcj}_{?}^{\mcz \rightarrow \mcx} \lbr \hat{\mcf}_{?}^{\mcz}, \mcv_{?}^{\mcz}, L_{glb}, L_{lcl} \rbr,\\
\end{aligned}\quad ? \in \lbe +, - \rbe.
\end{equation}
where $L_{glb}$ and $L_{lcl}$ are the global and local expansion level of the general conjugate Hilbert expansion strategies $\mcj_{?}^{\mcx \rightarrow \mcz}$ ($? \in \lbe +, - \rbe$). $\mcg$ is the multi-head group self-attention mechanism with $G$ group size, $\mce$ is the position embedding, and $\tilde{\mcj}_{?}^{\mcz \rightarrow \mcx}$ is the inverse expansion strategy of $\mcj_{?}^{\mcz \rightarrow \mcx}$.

\section{Experiments}
\label{sec:experiments}
\begin{table*}[t]
    \centering
    \vspace{-1em}
    \caption{The ablation study of different global levels of expansion strategies and different group sizes of SEGT encoder layer on the nuScenes validation set. 'T.L.', 'C.V.', 'Ped.', 'M.T.', 'Byc.', 'T.C.', and 'B.R.' are short for trailer, construction vehicle, pedestrian, motor, bicycle, traffic cone, and barrier, respectively.} 
    \label{tab:AblationofSEGT}
    \begin{adjustbox}{width=1\linewidth}
        \begin{tabular}{cc|cc|cccccccccc}
            \toprule
            $L_{glb}$ & $G$ & NDS $\uparrow$ & mAP $\uparrow$ & Car & Truck & Bus & T.L. & C.V. & Ped. & M.T. & Byc. & T.C. & B.R. \\
            \midrule
            5 &  90 & 71.8 & 67.2 & 87.8 & 62.7 & 76.6 & 43.9 & 27.0 & 88.8 & 75.0 & 61.2 & 78.2 & 70.9 \\
            5 & 128 & 71.9 & 67.2 & 88.1 & 61.9 & 76.6 & 42.9 & 26.4 & 88.6 & 75.6 & 61.7 & 78.4 & 72.3 \\
            5 & 256 & 72.1 & 67.6 & 88.0 & 63.7 & 78.6 & 45.0 & 28.3 & 88.7 & 75.2 & 60.4 & 78.0 & 70.1 \\
            6 &  90 & 71.8 & 67.4 & 87.9 & 63.0 & 76.4 & 45.2 & 26.3 & 88.8 & 76.3 & 62.0 & 77.2 & 70.9 \\
            6 & 128 & \textbf{72.5} & 68.1 & 88.4 & 65.2 & 76.2 & 47.1 & 26.8 & 89.1 & 77.0 & 61.1 & 78.3 & 71.3 \\
            6 & 256 & 72.1 & 67.4 & 87.9 & 62.9 & 78.2 & 45.3 & 26.1 & 88.8 & 76.4 & 58.3 & 79.3 & 71.1 \\
            7 &  90 & 72.1 & 67.8 & 88.4 & 63.3 & 77.0 & 46.6 & 27.2 & 88.9 & 75.4 & 60.7 & 78.2 & 72.1 \\
            7 & 128 & 72.2 & \textbf{68.2} & 88.2 & 65.2 & 77.3 & 47.5 & 28.9 & 89.0 & 76.2 & 60.8 & 78.0 & 70.6 \\
            7 & 256 & 72.2 & 68.0 & 88.1 & 64.1 & 78.5 & 47.4 & 28.4 & 88.8 & 76.8 & 61.2 & 78.2 & 68.1 \\
            \bottomrule
        \end{tabular}
    \end{adjustbox}
\end{table*}

\subsection{Dataset and Technical Details}

\paragraph{nuScenes Dataset.} The nuScenes \cite{Caesar2020nuScenes} dataset is a large-scale benchmark for 3D perception tasks in autonomous driving, comprising 1,000 driving sequences captured by 6 cameras, 5 radars, and 1 lidar with $360^{\circ}$ coverage. It includes detailed 3D bounding box annotations across 10 object categories with a long-tailed distribution, and is divided into training (700), validation (150), and testing (150) sequences. Evaluation is performed using the official metrics, including \textit{mean Average Precision (mAP)} and \textit{nuScenes detection score (NDS)}, with results reported according to the standard protocol, using 10 accumulated LiDAR scans as input.

\paragraph{Implementation Details.} For experiments on nuScenes, we apply four stages to the backbone of SEGT, and each stage with two SEGT encoder blocks. All the encoder block are 128 input channels. We set the detection range to $[-54.0~m, 54.0~m]$ for the X and Y axis, and $[-5~m, 3~m]$ for the Z axis. We use $[0.28125~m, 0.28125~m, 8.0~m]$ as the basic voxel size for experiments. 

\paragraph{Training and Inference.} All model variants are trained using the AdamW optimizer \cite{Loshchilov2019AdamW} on 8 NVIDIA A100 GPUs. The learning rate schedule follows the same configuration as described in \cite{Yin2021CenterPoint}. Inference times are evaluated on the same workstation (single NVIDIA A100 GPU), to ensure fair and comparable performance profiling.

\subsection{Benchmarks}

\mpara{3D object detection on nuScenes.} As shown in \mtab\ref{tab:ComparisonOnNuScenes}, our model outperforms all methods with remarkable gains. It achieves state-of-the-art performance, 73.9 and 70.1 in terms of test NDS and mAP without test-time augmentation (TTA), surpassing LION~\cite{Liu2024LION} by +0.0 and +0.3, respectively. Moreover, it achieves 74.5 and 71.2 in terms of test NDS and mAP with TTA, outperforming Real-Aug~\cite{Zhan2023RealAug} by +0.1 and +1.0, respectively. Notably, our method achieved the 1st place in the nuScenes lidar-based object detection task.

\subsection{Ablation Studies}

We conduct ablation studies to examine the effectiveness of each module of SEGT. The ablation study results are reported on nuScenes validation set. We study the effects of different global levels of general conjugate Hilbert expansion strategies $\mcj_{?}^{\mcx \rightarrow \mcz}$ ($? \in \lbe +, - \rbe$) and different group sizes of SEGT encoder layer, which introduced in \ref{tab:AblationofSEGT}. In view of the size of the detected objects themselves, appropriately increasing the global expansion level and the group size are beneficial for improving the mAP and NDS of SEGT on nuScenes.

\section{Conclusion}
\label{sec:conclusion}
In this paper, we present SEGT, a novel transformer framework for lidar-based 3D object detection tasks. To efficiently handle sparse point clouds, SEGT employs the group attention mechanism on voxels within distinct ordered fields, which are converted through a series of general spatial expansion strategies, to effectively enhance neighborhood representations. We validated our model on the nuScenes lidar-based object detection test dataset, achieving an NDS score of 73.9 without TTA, and 74.5 with TTA, demonstrating the effectiveness of our approach.

\newpage
{
    \small
    \bibliographystyle{unsrt}
    \bibliography{main}
}

\end{document}